\documentclass{bmvc2k}

\usepackage{amsmath}
\usepackage{amssymb}
\usepackage{algorithm}
\usepackage{algorithmic}
\usepackage{multirow}
\usepackage{graphicx} 
\usepackage{diagbox}
\usepackage{makecell}
\usepackage{comment}
\usepackage{hyperref}
\usepackage[misc]{ifsym}

\title{SamplingAug: On the Importance of Patch Sampling Augmentation for Single Image Super-Resolution}

\addauthor{Shizun Wang \textsuperscript{*}}{wangshizun@bupt.edu.cn}{1}
\addauthor{Ming Lu \textsuperscript{*}}{lu199192@gmail.com}{2}
\addauthor{Kaixin Chen}{2019140286@bupt.edu.cn}{1}
\addauthor{Jiaming Liu}{liujiaming@bupt.edu.cn}{1}
\addauthor{Xiaoqi Li}{xl3062@columbia.edu}{3}
\addauthor{Chuang zhang}{zhangchuang@bupt.edu.cn}{1}
\addauthor{Ming Wu \Letter}{wuming@bupt.edu.cn}{1}

\addinstitution{
 Beijing University of Posts and \\
 Telecommunications, China
}
\addinstitution{
 Intel Labs China
}
\addinstitution{
  Columbia university in the city \\
  of New York, USA
}

\runninghead{Wang, Lu, Chen, Liu, Li, Zhang, Wu}{SamplingAug}

\begin{document}

\maketitle

\begin{abstract}
With the development of Deep Neural Networks (DNNs), plenty of methods based on DNNs have been proposed for Single Image Super-Resolution (SISR). However, existing methods mostly train the DNNs on uniformly sampled LR-HR patch pairs, which makes them fail to fully exploit informative patches within the image. In this paper, we present a simple yet effective data augmentation method. We first devise a heuristic metric to evaluate the informative importance of each patch pair. In order to reduce the computational cost for all patch pairs, we further propose to optimize the calculation of our metric by integral image, achieving about two orders of magnitude speedup. The training patch pairs are sampled according to their informative importance with our method. Extensive experiments show our sampling augmentation can consistently improve the convergence and boost the performance of various SISR architectures, including EDSR, RCAN, RDN, SRCNN and ESPCN across different scaling factors ($\times 2$, $\times 3$, $\times 4$). Code is available at \url{https://github.com/littlepure2333/SamplingAug}.
\end{abstract}


\section{Introduction}
Single Image Super-Resolution (SISR) is a long-standing problem in computer vision. With the rapid development of Deep Neural Networks (DNNs) over the past few years, plenty of SISR methods based on DNNs are proposed. These methods mainly focus on network design \cite{shi2016real,dong2014learning,kim2016accurate,zhang2018image,lim2017enhanced}, real-world SR \cite{cai2019toward,xu2019towards}, zero-shot learning \cite{shocher2018zero,shaham2019singan}, meta learning \cite{park2020fast,soh2020meta}, network efficiency \cite{hui2018fast,xin2020binarized}, etc. However, as one of the most practical techniques, data augmentation has been rarely studied for SISR.

\begin{figure}[t]
\begin{center}
\begin{tabular}{cc}
\bmvaHangBox{\includegraphics[width=0.53\textwidth]{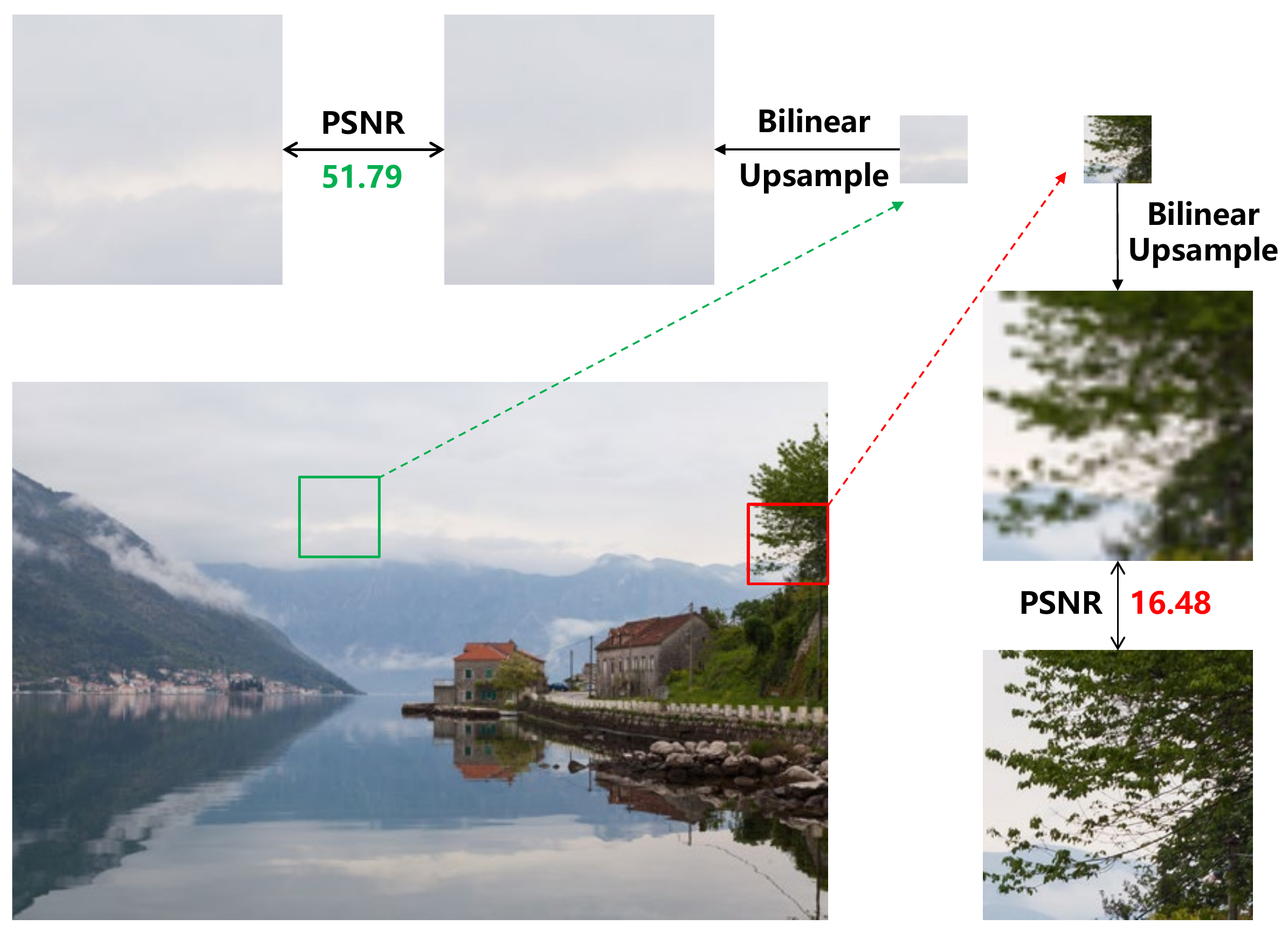}}&
\bmvaHangBox{\includegraphics[width=0.37\textwidth]{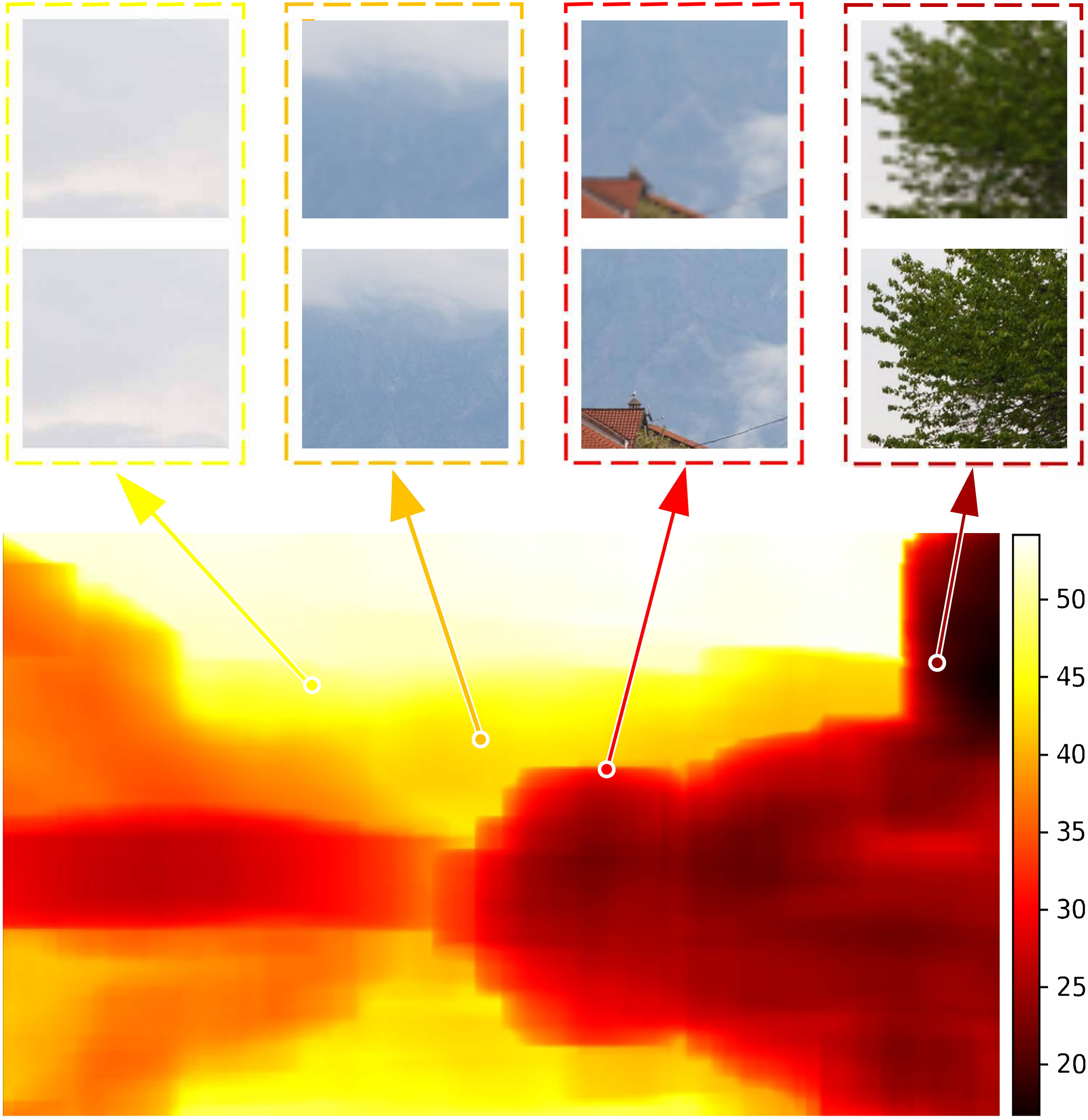}}\\
a) Illustration of informative importance & b) Illustration of patch sampling
\end{tabular}
\end{center}
\caption{\textbf{a) Illustration of informative importance.} There are both easy and hard patches within the image. Flat regions are easy to restore even with bilinear interpolation. However, patches with complicated texture are hard to restore. \textbf{b) Illustration of patch sampling.} According to the informative map, our method can sort all possible patches and take the informative patches for training.}
\label{fig:metric}
\end{figure}

Recently, \cite{yoo2020rethinking} provides a comprehensive analysis of applying data augmentation methods used in high-level vision tasks to SISR. Based on the analysis, they further propose a new augmentation method named CutBlur, which can reduce unrealistic distortions. Apart from CutBlur \cite{yoo2020rethinking}, former works mostly rely on simple geometric manipulations like rotation and flipping to the best of our knowledge. The motivation of CutBlur is to regularize a model to learn not only ``how'' but also ``where'' to apply the super-resolution to a given image. This indicates that spatial information is also essential for SISR data augmentation in addition to simple geometric manipulations.

Although CutBlur can achieve consistent improvements, it still follows the methodology of Mixup \cite{zhang2017mixup} and CutMix \cite{yun2019cutmix}, which are designed for high-level vision tasks. Since existing SISR methods mostly train the DNNs on sampled LR-HR patch pairs, we study the data augmentation problem from the perspective of patch sampling in this paper. Recent work on demosaicing and denoising \cite{sun2020learning} also investigates the problem of patch sampling. They train an additional PatchNet to select more informative patch samples for training and significantly enhance the network performance. In contrast to training an additional DNN, we propose a heuristic metric to evaluate the informative importance of each patch pair. Our metric is motivated by the fact that DNN can be treated as a highly non-linear function learned from data. Therefore, patches that can be super-resolved by linear function should be less informative for training the DNN. Inspired by this, we define the metric as the PSNR between linear SR and HR for each patch pair. 

However, directly calculating the metric for all patch pairs of an image is computationally complicated. We further propose to accelerate the calculation of our metric by integral image. Instead of using sliding windows, we first obtain the integral images of linear SR and HR separately, and then the integral images are used to deliver the metric for individual patch pair. By this way, we can significantly reduce the computational cost and achieve about two-orders of magnitude speedup.


After obtaining the informative importance of each patch pair, we need to sample a portion of patches for DNN training. We consider three sampling strategies. The first is simply selecting the $p\% $ most informative patches according to the metric. The second is applying Non-Maximum Suppression (NMS) \cite{girshick2015fast} to sampling the patches. The third is adopting Throwing-Dart (TD) sampling strategy \cite{gharbi2016deep} based on the metric. The latter two strategies can result in non-overlapped sampling across the image. All these three strategies can achieve consistent performance improvement on various popular architectures \cite{lim2017enhanced,zhang2018image,dong2014learning,shi2016real}. Besides, to our surprise, simply selecting the $p\% $ most informative patches achieves even slightly better results compared with NMS and TD, indicating that less informative samples might not contribute to the performance of SISR. We hope our paper can inspire future work on data augmentation for SISR.

Our contributions can be concluded as follows:
\begin{itemize}
\item We propose a simple heuristic metric, which can effectively measure the informative importance of each LR-HR patch pair for SISR training.  
\item We present an efficient method for metric calculation, which significantly reduces the computational cost.
\item We conduct extensive experiments with various SISR architectures, different scaling factors, and sampling strategies to demonstrate the benefit of our work.
\end{itemize}

\section{Related Work}
{\bf DNN-based Image Super-Resolution} Since the pioneering work SRCNN \cite{dong2014learning}, many DNN-based methods have been proposed. VDSR \cite{kim2016accurate} adopts a very deep DNN to learn the image residual. EDSR \cite{lim2017enhanced} analyzes the DNN layers and removes some redundant layers from SRResNet \cite{ledig2017photo}. RDN \cite{zhang2018residual} introduces dense connections to fully utilize the information of preceding layers. RCAN \cite{zhang2018image} explores the attentions mechanism for SR. FSRCNN \cite{dong2016accelerating} and ESPCN \cite{shi2016real} propose to use LR image as input and upscale the feature map at the end of DNNs. LAPAR \cite{li2020lapar} presents a method based on linearly-assembled pixel-adaptive regression network. ClassSR \cite{kong2021classsr} accelerates SR networks on large images (2K-8K) by combining classification and SR in a unified framework. Although plenty of DNN-based methods are proposed, as one of the most practical ways to improve model performance, data augmentation has been rarely studied for SISR.


{\bf Patch Sampling} To the best of our knowledge, DNNs of SR are mostly trained on uniformly sampled LR-HR patch pairs and tested on images. However, there are hard and simple areas within a single image. KPN \cite{bako2017kernel} uses a sampling strategy named Throwing Dart (TD), according to the importance map delivered from additional rendered buffers. PatchNet \cite{sun2020learning} recently learns to select the most useful patches from an image to construct a training set instead of random selection. They significantly boost the performance of demosaicing and denoising. Our work also tries to select the informative patch pairs for DNN training. In contrast to training a network, we propose an efficient heuristic metric to evaluate the informative importance of LR-HR patch pair.

\section{Method}
In this part, we first present the effective metric that evaluates the informative importance of patch pair in Section \ref{sec:metric}. Then we introduce an acceleration solution based on integral image in Section \ref{sec:integral}. Finally, we briefly describe three sampling strategies in Section \ref{sec:sampling}.

\subsection{Informative Importance Metric}
\label{sec:metric}

Existing SISR methods \cite{lim2017enhanced,zhang2018image} mostly train the DNNs with sampled LR-HR patch pairs rather than the whole images due to the memory limitation. The patches are usually uniformly sampled to construct a training batch. As shown in Fig \ref{fig:metric} a), we observe that there are easy patches and hard patches even within a single image. For example, flat regions are easy to restore even with bilinear interpolation, while regions with complicated textures are hard to restore. Furthermore, the recent success of DNN is largely due to its non-linear capability learned from database. Inspired by this observation, we propose to explore the informative importance of each patch pair by defining a heuristic metric. To be more specific, let $x^{LR}$ and $x^{HR}$ be the LR and HR patches. The patch size of $x^{HR}$ is $k \times k$ and the scaling factor is $s$. We first upscale $x^{LR}$ to the SR patch ${x^{\widetilde {SR}}}$ with bilinear interpolation. Then we define the heuristic metric as the PSNR between ${x^{\widetilde {SR}}}$ and $x^{HR}$. This process can be formulated as follow:

\begin{equation}
MSE = \frac{1}{k^2} \sum_{i=0}^{k-1}\sum_{j=0}^{k-1} ||x_{i,j}^{\widetilde {SR}} - x_{i,j}^{HR}||_2^2
\label{eq:mse}
\end{equation}

\begin{equation}
PSNR = 10 * log_{10}(\frac{MAX_x^2}{MSE})
\end{equation}

where $MAX_x$ is the maximum possible pixel value of the image and is set to 255 for 8-bit format. {\bf Higher PSNR between ${x^{\widetilde {SR}}}$ and $x^{HR}$ indicates that this patch is less informative for DNN training since the linear restoration can already achieve pleasing restoration results.} As illustrated by Fig \ref{fig:metric} b), for every image in the database, we can sort all the possible patch pairs by calculating the PSNR between ${x^{\widetilde {SR}}}$ and $x^{HR}$, and take the $p\% $ most informative patches for training.

\subsection{Integral Image Acceleration}
\label{sec:integral}

As described previously, our method needs to calculate the PSNR between all possible patch pairs. Assume there are $N$ images under resolution $H \times W$, the computational complexity of sliding windows is $O(NHW)$. Although it can be accelerated by GPU parallel computing, our primary GPU implementation still takes about 10 seconds to process an image in 2K resolution.

To overcome this, we optimize the calculation process using integral image \cite{crow1984summed}. Specifically, instead of bilinearly interpolating the local patch, we upscale the LR image ${X^{LR}}$ to ${X^{\widetilde {SR}}}$. For a patch pair (top left pixel location is $(u,v)$), we can represent the MSE in Eq. \ref{eq:mse} as:

\begin{equation}
\begin{aligned}
MSE = \frac{1}{{{k^2}}}\sum\limits_{i = 0}^{k - 1} {\sum\limits_{j = 0}^{k - 1} {\left\| {X_{(u + i,j + v)}^{\widetilde {SR}} - X_{(u + i,j + v)}^{HR}} \right\|_2^2} }
\end{aligned}
\end{equation}

The above equation can be expanded and summarized into:

\begin{equation}
\begin{aligned}
MSE = \frac{1}{{{k^2}}}\sum\limits_{i = 0}^{k - 1} {\sum\limits_{j = 0}^{k - 1} {X_{(u + i,j + v)}^{HR}X_{(u + i,j + v)}^{HR}} }  
+ \frac{1}{{{k^2}}}\sum\limits_{i = 0}^{k - 1} {\sum\limits_{j = 0}^{k - 1} {X_{(u + i,j + v)}^{\widetilde {SR}}X_{(u + i,j + v)}^{\widetilde {SR}}} }  \\
- 2\frac{1}{{{k^2}}}\sum\limits_{i = 0}^{k - 1} {\sum\limits_{j = 0}^{k - 1} {X_{(u + i,j + v)}^{HR}X_{(u + i,j + v)}^{\widetilde {SR}}} } 
\end{aligned}
\label{eq:sum}
\end{equation}

For each term in Eq. \ref{eq:sum}, we observe it can be significantly accelerated by integral image. We denote the integral image as $\gamma ( * )$ and reformulate the first term as:

\begin{equation}
\begin{aligned}
\sum\limits_{i = 0}^{k - 1} {\sum\limits_{j = 0}^{k - 1} {X_{(u + i,j + v)}^{HR}X_{(u + i,j + v)}^{HR}} }  = 
\gamma {({X^{HR}} \odot {X^{HR}})_{\overline u,\overline v}} + \gamma {({X^{HR}} \odot {X^{HR}})_{u,v}} \\- \gamma {({X^{HR}} \odot {X^{HR}})_{\overline u,v}} - \gamma {({X^{HR}} \odot {X^{HR}})_{u,\overline v}}
\end{aligned}
\label{eq:int1}
\end{equation}

where $ \odot $ denotes the element-wise multiplication, $\overline u  = u + k - 1$ and $\overline v  = v + k - 1$ are the bottom right pixel locations. We omit the constant $\frac{1}{{{k^2}}}$ for simplicity. The second and third terms of Eq. \ref{eq:sum} can be reformulated similarly. Using integral image \cite{crow1984summed} avoids traversing all patch pairs, greatly reducing the computational complexity. The proposed algorithm takes 0.1 second to process a 2K image, achieving about two orders of magnitude speedup compared with primary GPU implementation (10 seconds). 

\begin{figure}
\begin{center}
\begin{tabular}{ccc}
\bmvaHangBox{\includegraphics[width=0.3\textwidth]{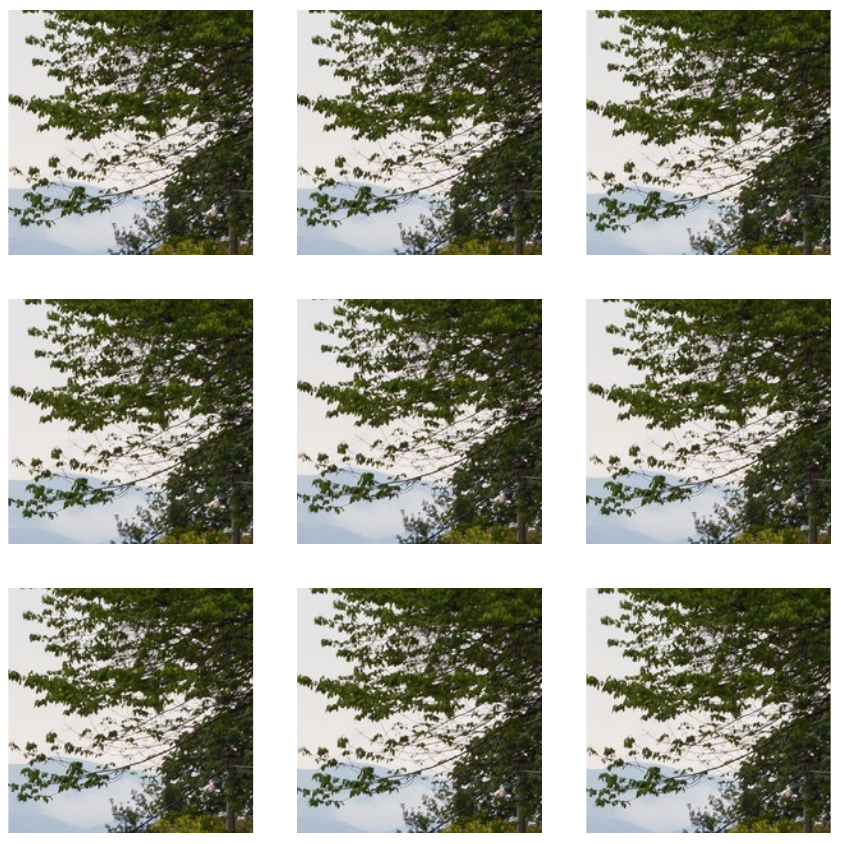}}&
\bmvaHangBox{\includegraphics[width=0.3\textwidth]{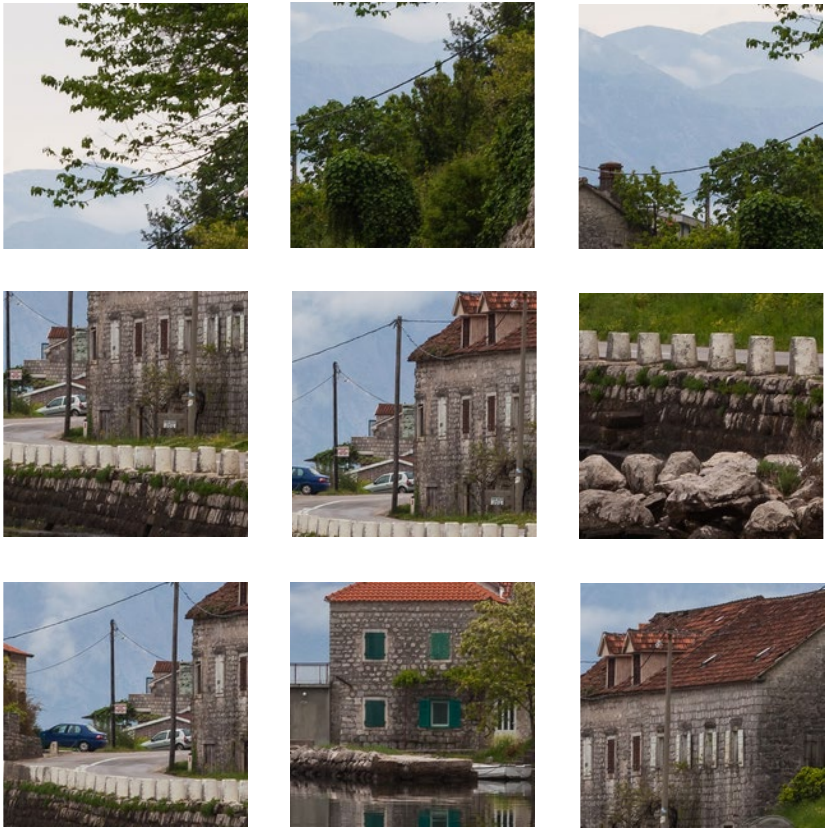}}&
\bmvaHangBox{\includegraphics[width=0.3\textwidth]{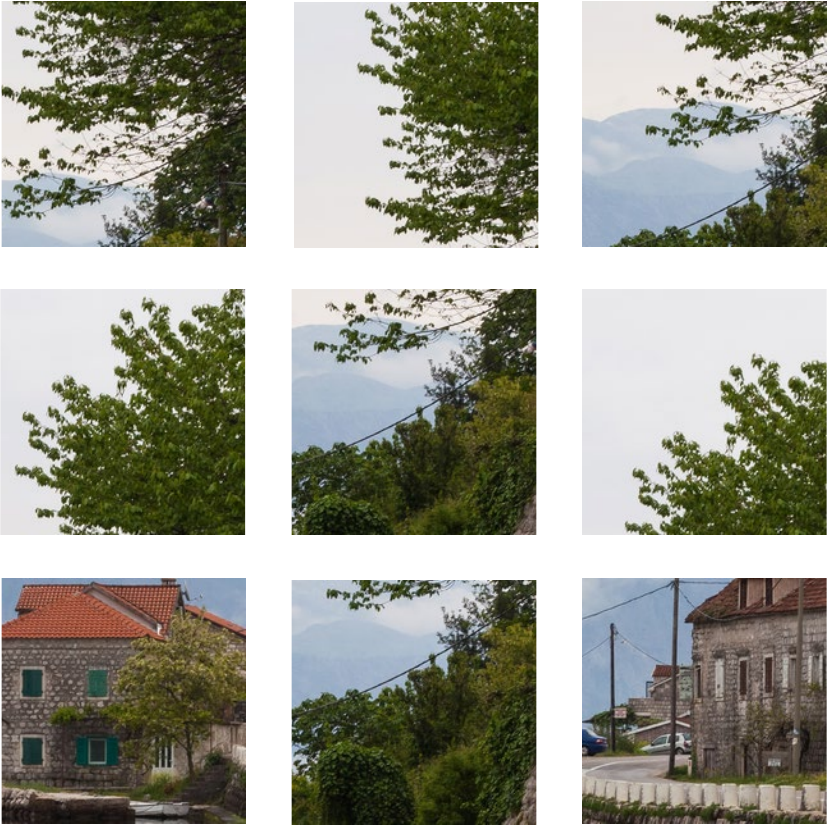}}\\
\small{a) GS} & \small{b) TDS} & \small{c) NMSS}
\end{tabular}
\end{center}
   \caption{Example of selected patches with different sampling strategies. Greedy sampling will select overlapped patches, while Throwing Dart sampling and NMS sampling result in non-overlapped patches.}
\label{fig:hard-patch-gather}
\end{figure}

\subsection{Different Sampling Strategies}
\label{sec:sampling}

{\bf Greedy Sampling (GS)} Based on the informative importance map, we can use greedy sampling strategy, which directly gathers the $p\% $ most informative patches. As shown in Fig. \ref{fig:hard-patch-gather}, greedy sampling will result in heavily overlapped patches. {\bf This sampling strategy is our default strategy used in the paper.} To further study the benefit of non-overlapped sampling, we also design two variant strategies.

{\bf Throwing Dart Sampling (TDS)} Throwing-Dart Sampling \cite{bako2017kernel} can automatically choose non-overlapped patches. This strategy first generates non-overlapped candidates by employing throwing darts. The patch candidates are pruned according to the importance map to deliver $p\% $ patches for training. \cite{bako2017kernel} adopts the rendered normal and color images to obtain the importance map. In this work, we use the proposed informative metric to obtain the importance map. As illustrated by Fig. \ref{fig:hard-patch-gather}, TDS can construct non-overlapped patches for DNN training.

{\bf Non-maximum Suppression Sampling (NMSS)}
Non-maximum suppression \cite{girshick2015fast} is a technique widely used in object detection. Inspired by NMS, we first select the most informative patch and mask out all overlapped patches with IoU higher than a threshold. Then, we select the next most informative patch from the remaining candidate patches. This process is repeated to construct $p\% $ patches as the training samples. In this manner, NMSS will select non-overlapped patches for training. We show the examples of selected patches in Fig. \ref{fig:hard-patch-gather}.  

\section{Experiment}
\subsection{Experimental setup}
We use DIV2K dataset \cite{Agustsson_2017_CVPR_Workshops} for training, which consists of 1000 high-resolution images in 2K resolution. The low-resolution images are generated by bicubic downsampling with scaling factors of $\times 2$, $\times 3$ and $\times 4$. Following former works, we use 800 images for training and 10 images for validation. We also report results on four standard benchmark datasets including Set5, Set14, B100 and Urban100. Peak Signal-to-Noise Ratio (PSNR) and Structural Similarity (SSIM) are adopted as the evaluation metrics to measure SR performance. We ignore image borders and calculate the PSNR and SSIM in the luminance channels.

\subsection{Implementation Details}
To verify the effectiveness and generalization of our method, we use 5 popular SISR architectures as baselines including SRCNN, ESPCN, EDSR, RCAN and RDN. All the experiments are conducted with PyTorch framework on NVIDIA 2080Ti GPUs. The patch size is set to $192 \times 192$ during training unless mentioned otherwise. We use Adam optimizer with $\beta_1 = 0.9$ and $\beta_2 = 0.999$. All models are trained 300 epochs from scratch using the above experimental setups. By integral Image Acceleration, processing whole DIV2K dataset takes additional 3 minutes for greedy sampling, 5 minutes for NMS sampling and 1 minutes for throwing darts sampling. The additional cost of our method is almost negligible.

\subsection{Results on Various SISR Architectures}

\subsubsection{Quantitative Results}
In Tab. \ref{tab:bench}, we show a comprehensive quantitative comparison with various baseline models under different scaling factors. As can be seen, the proposed method consistently boosts the PSNR performance across various popular baseline architectures, scaling factors and benchmark datasets by adequate margins. We also provide the SSIM results in the supplementary material. Surprisingly, for the lightweight models, our method brings significant benefits. For example, on Set5 and scaling factor $\times 2$, SRCNN and ESPCN obtain huge margins of \textbf{1.40 dB} and \textbf{2.26 dB} respectively. All the results show that our method can yield a consistent and significant performance boost.

\begin{table}[t]
\begin{center}
\scriptsize
\begin{tabular}{|c|c|c|c|c|c|}
\hline
Networks    & scale      & Set5            & Set14           & B100            & Urban100   \\ \hline\hline
            &x2          &34.94 (+1.40)    &31.62 (+1.11)    &30.47 (+0.87)    &27.79 (+0.90) \\
SRCNN       &x3          &31.37 (+0.47)    &28.60 (+0.40)    &27.77 (+0.26)    &25.06 (+0.28) \\
            &x4          &29.33 (+0.35)    &26.91 (+0.30)    &26.42 (+0.16)    &23.63 (+0.18) \\\hline
            &x2          &34.05 (+2.26)    &31.14 (+1.91)    &30.36 (+1.23)    &27.64 (+1.30) \\
ESPCN       &x3          &30.36 (+1.36)    &27.93 (+1.07)    &27.50 (+0.70)    &24.76 (+0.75) \\
            &x4          &28.20 (+1.26)    &26.13 (+0.89)    &26.10 (+0.65)    &23.28 (+0.68) \\\hline
            &x2          &37.27 (+0.40)    &33.38 (+0.41)    &31.68 (+0.27)    &30.46 (+0.86) \\
EDSR        &x3          &33.02 (+0.33)    &29.86 (+0.25)    &28.49 (+0.13)    &26.44 (+0.35) \\
            &x4          &30.66 (+0.33)    &27.93 (+0.30)    &26.94 (+0.11)    &24.58 (+0.27) \\\hline
            &x2          &36.85 (+0.15)    &32.97 (+0.09)    &31.40 (+0.06)    &29.62 (+0.21) \\
RCAN        &x3          &32.71 (+0.58)    &29.61 (+0.39)    &28.34 (+0.22)    &26.16 (+0.50) \\
            &x4          &30.35 (+0.48)    &27.68 (+0.42)    &26.83 (+0.18)    &24.37 (+0.36) \\\hline
            &x2          &37.20 (+0.07)    &33.25 (+0.17)    &31.65 (+0.11)    &30.22 (+0.38) \\
RDN         &x3          &33.15 (+0.34)    &29.87 (+0.18)    &28.57 (+0.15)    &26.51 (+0.36) \\
            &x4          &30.70 (+0.34)    &28.00 (+0.33)    &26.99 (+0.14)    &24.63 (+0.31) \\\hline
\end{tabular}
\end{center}
\caption{Quantitative Results (PSNR) on Benchmarks. We omit the results of baselines for simplicity, and only report our results and the gain margins. SSIM results are reported in supplemental material.}
\label{tab:bench}
\end{table}

\vspace*{-5mm}
\begin{figure}[t]
\begin{center}
\begin{tabular}{cc}
\bmvaHangBox{\includegraphics[width=0.48\textwidth]{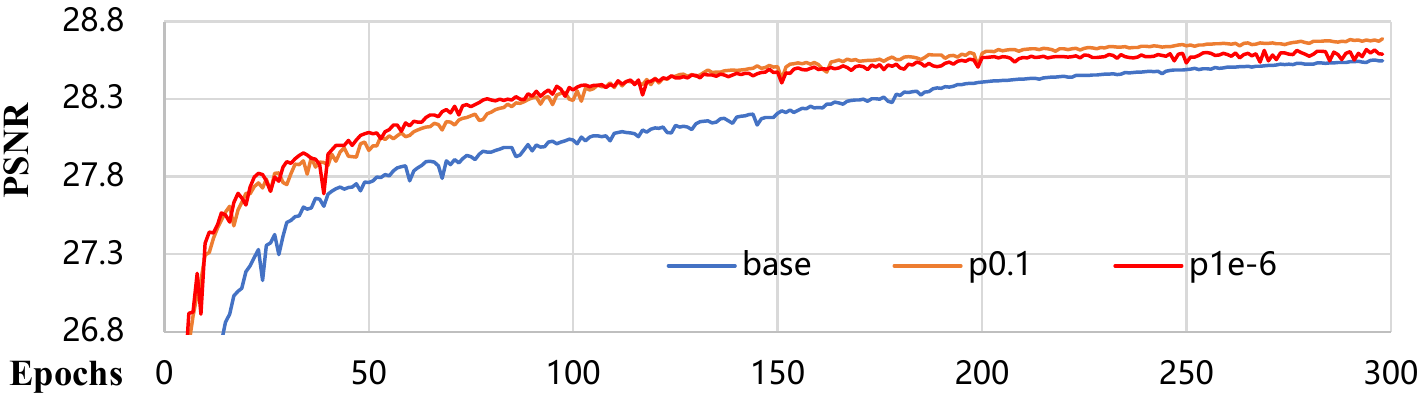}}&
\bmvaHangBox{\includegraphics[width=0.48\textwidth]{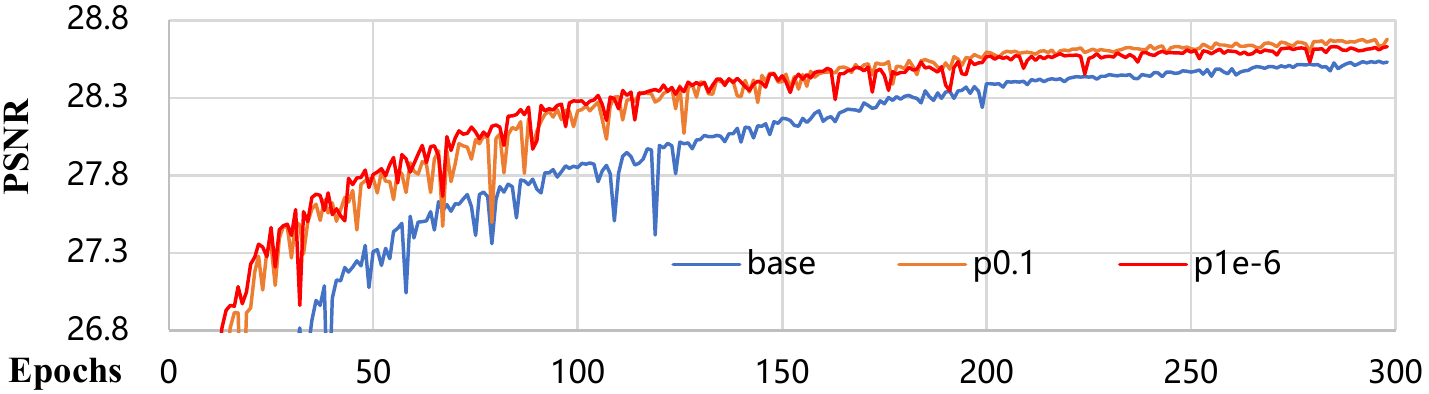}}\\
\small{a) EDSR} & \small{b) RDN}\\
\end{tabular}
\end{center}
\vspace*{-1mm}
\caption{Convergence analysis of our method. We show the training process of EDSR and RDN in scaling factor $\times 4$. Our results of $p = 0.1$ and $p = 1e-6$ are shown in this figure for illustration. As can be seen, our method can consistently improve the convergence and performance of SR networks.}
\label{fig:convergence}
\end{figure}

\vspace*{-2mm}
\subsubsection{Convergence Results}
Fig. \ref{fig:convergence} shows the convergence comparison between the baselines and our method. It can be observed that our method can also help the training process converge faster. In particular, the results of our method at 150 epochs already achieve the performance of baselines at 300 epochs. Fig. \ref{fig:convergence} also shows our results of different sampling portion $p\% $. We observe that selecting fewer patches ($p = 1e - 6$) increases the convergence speed at the early stage, while more informative patches ($p = 0.1$) can achieve better results after convergence. This enables our method to achieve a flexible balance between training time and performance.

\vspace*{-2mm}
\subsubsection{Qualitative Results}
\begin{figure}[t]
\begin{center}
\includegraphics[width=\textwidth]{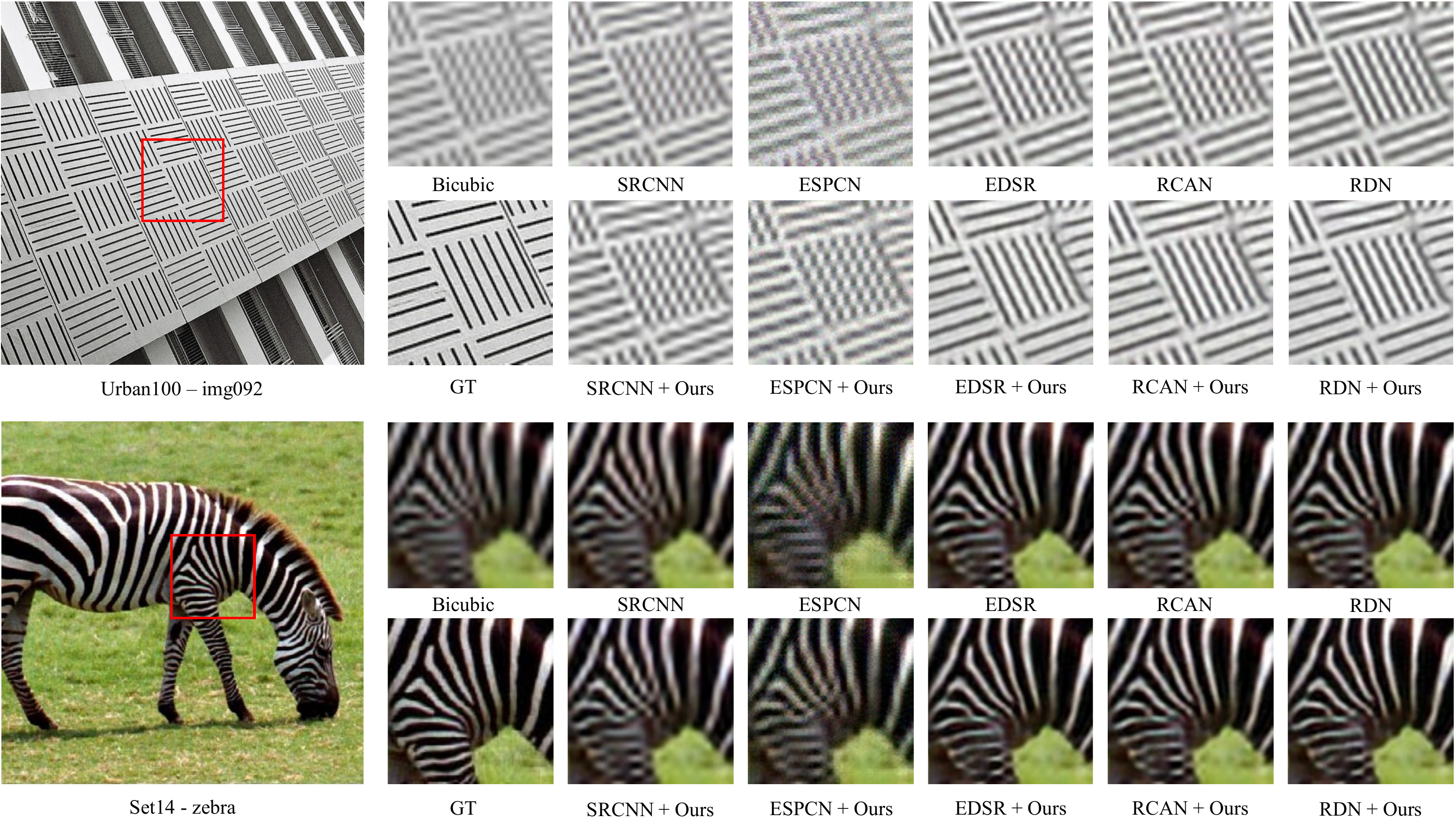}
\end{center}
   \caption{Qualitative comparison between our method and baselines (x4 SR). More results can be found in the supplementary material.}
\label{fig:visual}
\end{figure}

Fig. \ref{fig:visual} shows the qualitative results of our method against the baselines. All the SR networks benefit from our method and obtain visually better results. We provide more qualitative results in the supplementary materials.

\subsection{Ablation Study}

\subsubsection{Impact of patch size.}
Since our method measures the informative importance for each LR-HR patch pair, it is essential to study the impact of patch size to evaluate our method. We show the results of our method with different patch sizes in Tab. \ref{tab:patch-size}. Our method obtains consistent and significant gain margins, and especially effective for smaller patch size.

\begin{table}[]
\begin{minipage}[t]{0.38\linewidth}
\vspace{0pt}
\centering
\scriptsize
\begin{tabular}{|l|l|l|}
\hline
patch size   & baseline   & ours                 \\ \hline\hline
32           & 27.71      & 28.27 (+0.56)        \\ \hline
64           & 28.10      & 28.48 (+0.38)        \\ \hline
128          & 28.16      & 28.50 (+0.34)        \\ \hline
192          & 28.55      & 28.69 (+0.14)        \\ \hline
256          & 28.58      & 28.69 (+0.11)        \\ \hline
384          & 28.90      & 28.95 (+0.05)        \\ \hline
\end{tabular}
\vspace*{+3mm}
\caption{Ablation study on different patch size. The PSNR results of EDSR are reported in DIV2K dataset at x4 scale.}
\label{tab:patch-size}
\end{minipage}\hspace{0.02\linewidth}
\begin{minipage}[t]{0.6\linewidth}
\vspace{0pt}
\centering
\scriptsize
\begin{tabular}{|c|l|l|}
\hline
\diagbox {portion}{model} & ESPCN                  & EDSR                   \\ \hline\hline
baseline                  & 29.24                  & 34.56                  \\ \hline
0.5                       & 31.04 (+1.80)          & 34.67 (+0.11)          \\ \hline
0.3                       & 31.34 (+2.10)          & 34.68 (+0.12)          \\ \hline
0.1                       & 31.52 (+2.28)          & \textbf{34.68 (+0.12)} \\ \hline
1e-2                      & 31.72 (+2.48)          & 34.64 (+0.08)          \\ \hline
1e-3                      & 31.78 (+2.54)          & 34.62 (+0.06)          \\ \hline
1e-4                      & 31.77 (+2.53)          & 34.61 (+0.05)          \\ \hline
1e-5                      & 31.78 (+2.54)          & 34.60 (+0.04)          \\ \hline
1e-6                      & \textbf{31.78 (+2.54)} & 34.62 (+0.06)          \\ \hline
\end{tabular}
\vspace*{+2mm}
\caption{Ablation study of varying portion. The PSNR results are reported in DIV2K dataset at x2 scale.}
\label{tab:vary-p}
\end{minipage}
\end{table}

\subsubsection{Impact of sampled portion.}
\label{exp:p}
We conduct extensive experiments to explore the SR performance under different portions of informative data. The results in Tab. \ref{tab:vary-p} show that networks with strong capacity (EDSR, RCAN and RDN) obtain the best performance when $p$ is set to 0.1. We think 10\% most informative patch pairs are the optimal choice for these networks, since more data will introduce ``noisy'' samples and less data cannot meet the capacity of these networks. While for lightweight networks like SRCNN and ESPCN, they achieve the best performance when $p$ is set to $1e^{-5}$, this is perhaps because their capacity is less strong and a few of most informative samples already meet their capacity. Therefore, stronger capacity models can utilize more informative data to achieve the best performance. Nevertheless, even when $p$ is extremely small, the proposed method can still boost the performance compared with baselines. To our surprise, just only selecting one most informative patch per image ($p=1e^{-6}$) can already overpass the results of uniformly sampling all patches, validating the effectiveness of our method.

\subsubsection{Do other metrics help?}
Apart from our metric, we also try other heuristic metrics to measure the informative importance of local patch. We use the Standard Deviation (STD) of the patch since STD can kind of model the texture information. We also try to use edge detection operators (Sobel, Canny and Laplacian) to measure the informative importance. Tab. \ref{tab:psnr-std-loss} shows the results of these metrics, where $std0$ means the STD of all channels (RGB), $std1$ means the mean of each channel’s STD (RGB), and $std2$ means the STD of y channel (YCbCr). As can be seen, our method achieves better results compared with these metrics, demonstrating the advantages of our metric. All the results show the effectiveness of our metric to measure the informative importance of LR-HR patch pair.

\subsubsection{Do sampling strategies matter?}
As mentioned above, we simply adopt the Greedy Sampling (GS) strategy, which will sample heavily overlapped patches for training. In order to study the benefit of non-overlapped sampling, we compare GS with NMS \cite{girshick2015fast} sampling and Throwing-Dart sampling \cite{bako2017kernel}. The results are reported in Tab. \ref{tab:darts-nms}. In this experiment, we adopt the three sampling strategies to construct the same number ($1e^4$) patches per image for training. The results in Tab. \ref{tab:darts-nms} show that sampling non-overlapped patches bring little benefit compared with greedy sampling. This is because our metric can successfully measure the informative importance and less informative samples contribute little to the final performance.

\begin{table}[]
\begin{minipage}[t]{0.48\linewidth} 
\vspace{0pt}
\centering
\scriptsize
\begin{tabular}{|c|c|c|}
\hline
metrics           & PSNR           & gain                     \\ \hline\hline
baseline          & 34.56          &                          \\ \hline
std0              & 34.57          & +0.01                    \\ \hline
std1              & 34.58          & +0.02                    \\ \hline
std2              & 34.58          & +0.02                    \\ \hline
Sobel             & 34.66          & +0.10                    \\ \hline
Canny             & 34.65          & +0.09                    \\ \hline
Laplacian         & 34.63          & +0.07                    \\ \hline
\textbf{Ours}     & \textbf{34.68} & \textbf{+0.12}           \\ \hline
\end{tabular}
\vspace*{+2mm}
\caption{Ablation study of other metrics. The PSNR results of EDSR on DIV2K at x2 scaling factor are reported.}
\vspace*{-2mm}
\label{tab:psnr-std-loss}
\end{minipage}\hspace{0.01\linewidth}
\begin{minipage}[t]{0.48\linewidth}
\vspace{0pt}
\centering
\scriptsize
\begin{tabular}{|c|c|l|}
\hline
backbone               & sampling method  & PSNR            \\ \hline\hline
\multirow{4}{*}{EDSR}  & baseline         & 34.56                  \\ \cline{2-3} 
                       & TDS              & 34.67 (+0.11)          \\ 
                       & NMSS             & \textbf{34.69 (+0.13)} \\ 
                       & GS               & 34.68 (+0.12)          \\ \hline
\multirow{4}{*}{RCAN}  & baseline         & 34.45                  \\ \cline{2-3} 
                       & TDS              & 34.54 (+0.09)          \\ 
                       & NMSS             & 34.54 (+0.09)          \\ 
                       & GS               & \textbf{34.55 (+0.10)} \\ \hline
                        
\end{tabular}
\vspace*{+2mm}
\caption{The PSNR results of different sampling strategies on DIV2K at x2 scaling factor.}
\vspace*{-2mm}
\label{tab:darts-nms}
\end{minipage}
\end{table}


\subsection{Comparison with Other methods}
\begin{figure*}[t]
\begin{center}
\bmvaHangBox{\includegraphics[width=\textwidth]{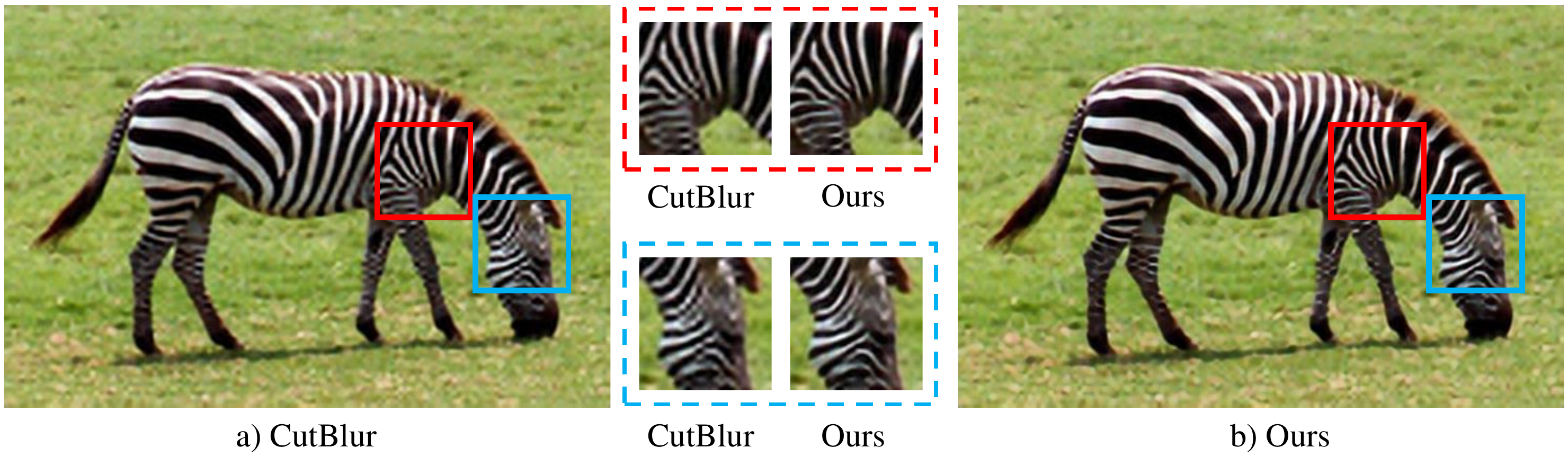}}
\end{center}
  \caption{Qualitative results of our method against CutBlur \cite{yoo2020rethinking}.}
\label{fig:cutblur}
\end{figure*}

CutBlur \cite{yoo2020rethinking} is a data augmentation method designed for SR. We compare our method with CutBlur. The comparison results between CutBlur and our method are shown in Tab. \ref{tab:cutblur}. Without pretraining, CutBlur obtains even worse results compared with baselines, while our method can still improve the performance. With pretraining, our method can achieve larger gain margin compared with CutBlur. Our method focuses on the sampling aspect of SR, while CutBlur still follows the methodology of data augmentation methods used in high-level vision tasks. This comparison shows the advantages of our method over CutBlur.The qualitative results can be found in Fig. \ref{fig:cutblur}.

\begin{table}[t]
\begin{minipage}[t]{0.46\linewidth} 
\vspace{0pt}
\begin{center}
\scriptsize
\begin{tabular}{|c|c|c|}
\hline
method          & PSNR           & gain                     \\ \hline\hline
Baseline        & 34.56          &                          \\ \hline
ours 0.1        & \textbf{34.68} & \textbf{+0.12}           \\ \hline
ours 0.3        & 34.68          & +0.12                    \\ \hline
ours 0.5        & 34.67          & +0.11                    \\ \hline
PatchNet        & 34.65          & +0.09                    \\ \hline
OHEM 0.1        & 34.39          & \textcolor{green}{-0.17} \\ \hline
OHEM 0.3        & 34.56          & \textcolor{green}{-0.01} \\ \hline
OHEM 0.5        & 34.62          & +0.06                    \\ \hline

\end{tabular}
\end{center}
\caption{Comparison with PatchNet and Online Hard example Mining (OHEM). The PSNR results of EDSR are reported in DIV2K dataset at x2 scale. The decimal indicates the portion of dataset.}
\label{tab:methods-online}
\end{minipage}\hspace{0.02\linewidth}
\begin{minipage}[t]{0.48\linewidth}
\vspace{0pt}
\begin{center}
\scriptsize
\begin{tabular}{|l|l|l|l|}
\hline
\multicolumn{2}{|c|}{Ours}	              & \multicolumn{2}{c|}{CutBlur}                      \\ \hline
portion	       & PSNR	          & portion      & PSNR                        \\ \hline\hline
baseline       & 28.55	                  & baseline     & 28.56                              \\ \hline
0.5	           & 28.68 (+0.13)	          & 0.9          & 28.29 \textcolor{green}{(-0.27)}   \\ 
0.1            & 28.69 (+0.14)	          & 0.5          & 28.39 \textcolor{green}{(-0.17)}   \\ 
1e-6	       & 28.62 (+0.07)	          & 0.1          & 28.37 \textcolor{green}{(-0.19)}   \\ \hline
0.5 \ \ + pre  & \textbf{28.93 (+0.38)}   & 0.9 + pre    & 28.73 (+0.17)                      \\ 
0.1 \ \ + pre  & 28.90 (+0.35)            & 0.5 + pre    & 28.74 (+0.18)                      \\ 
1e-6 + pre     & 28.79 (+0.24)	          & 0.1 + pre    & 28.73 (+0.17)                      \\ \hline
\end{tabular}
\end{center}
\caption{Comparison of our method against CutBlur \cite{yoo2020rethinking}. The PSNR results of EDSR are reported in DIV2K dataset at x4 scale. $pre$ denotes pretraining.}
\label{tab:cutblur}
\end{minipage}
\end{table}

We also compare our approach against two online sampling methods PatchNet \cite{sun2020learning} and OHEM \cite{shrivastava2016training}. 
PatchNet trains an additional network to measure the trainability of each patch, and only uses the trainable patches during training. OHEM only back-propagates the highest losses in every training batch. We observe that OHEM is unstable to converge in SR training. As can be seen in Tab. \ref{tab:methods-online}, hard negative mining fails to achieve consistent performance improvement and our approach achieves better results compared with PatchNet. All the results show the effectiveness of our method.

\section{Conclusion}
In this paper, we study the data augmentation problem for SR in terms of patch sampling. we first devise a simple yet effective metric to measure the informative importance of LR-HR patch pairs. Then we propose an efficient solution based on integral image, which significantly reduces the computational cost of calculating this metric for every image. Instead of uniformly sampling all patches for SR training, we sample the most informative patches. Extensive experiments are conducted to validate the effectiveness and advantages of the proposed method across various architectures, scaling factors and datasets. We hope this paper can inspire future works on data augmentation for SR.

\bibliography{egbib}

\begin{thebibliography}{28}
\providecommand{\natexlab}[1]{#1}
\providecommand{\url}[1]{\texttt{#1}}
\expandafter\ifx\csname urlstyle\endcsname\relax
  \providecommand{\doi}[1]{doi: #1}\else
  \providecommand{\doi}{doi: \begingroup \urlstyle{rm}\Url}\fi

\bibitem[Agustsson and Timofte(2017)]{Agustsson_2017_CVPR_Workshops}
Eirikur Agustsson and Radu Timofte.
\newblock Ntire 2017 challenge on single image super-resolution: Dataset and
  study.
\newblock In \emph{The IEEE Conference on Computer Vision and Pattern
  Recognition (CVPR) Workshops}, July 2017.

\bibitem[Bako et~al.(2017)Bako, Vogels, McWilliams, Meyer, Nov{\'a}k, Harvill,
  Sen, Derose, and Rousselle]{bako2017kernel}
Steve Bako, Thijs Vogels, Brian McWilliams, Mark Meyer, Jan Nov{\'a}k, Alex
  Harvill, Pradeep Sen, Tony Derose, and Fabrice Rousselle.
\newblock Kernel-predicting convolutional networks for denoising monte carlo
  renderings.
\newblock \emph{ACM Trans. Graph.}, 36\penalty0 (4):\penalty0 97--1, 2017.

\bibitem[Cai et~al.(2019)Cai, Zeng, Yong, Cao, and Zhang]{cai2019toward}
Jianrui Cai, Hui Zeng, Hongwei Yong, Zisheng Cao, and Lei Zhang.
\newblock Toward real-world single image super-resolution: A new benchmark and
  a new model.
\newblock In \emph{Proceedings of the IEEE/CVF International Conference on
  Computer Vision}, pages 3086--3095, 2019.

\bibitem[Crow(1984)]{crow1984summed}
Franklin~C Crow.
\newblock Summed-area tables for texture mapping.
\newblock In \emph{Proceedings of the 11th annual conference on Computer
  graphics and interactive techniques}, pages 207--212, 1984.

\bibitem[Dong et~al.(2014)Dong, Loy, He, and Tang]{dong2014learning}
Chao Dong, Chen~Change Loy, Kaiming He, and Xiaoou Tang.
\newblock Learning a deep convolutional network for image super-resolution.
\newblock In \emph{European conference on computer vision}, pages 184--199.
  Springer, 2014.

\bibitem[Dong et~al.(2016)Dong, Loy, and Tang]{dong2016accelerating}
Chao Dong, Chen~Change Loy, and Xiaoou Tang.
\newblock Accelerating the super-resolution convolutional neural network.
\newblock In \emph{European conference on computer vision}, pages 391--407.
  Springer, 2016.

\bibitem[Gharbi et~al.(2016)Gharbi, Chaurasia, Paris, and
  Durand]{gharbi2016deep}
Micha{\"e}l Gharbi, Gaurav Chaurasia, Sylvain Paris, and Fr{\'e}do Durand.
\newblock Deep joint demosaicking and denoising.
\newblock \emph{ACM Transactions on Graphics (TOG)}, 35\penalty0 (6):\penalty0
  1--12, 2016.

\bibitem[Girshick(2015)]{girshick2015fast}
Ross Girshick.
\newblock Fast r-cnn.
\newblock In \emph{Proceedings of the IEEE international conference on computer
  vision}, pages 1440--1448, 2015.

\bibitem[Hui et~al.(2018)Hui, Wang, and Gao]{hui2018fast}
Zheng Hui, Xiumei Wang, and Xinbo Gao.
\newblock Fast and accurate single image super-resolution via information
  distillation network.
\newblock In \emph{Proceedings of the IEEE conference on computer vision and
  pattern recognition}, pages 723--731, 2018.

\bibitem[Kim et~al.(2016)Kim, Kwon~Lee, and Mu~Lee]{kim2016accurate}
Jiwon Kim, Jung Kwon~Lee, and Kyoung Mu~Lee.
\newblock Accurate image super-resolution using very deep convolutional
  networks.
\newblock In \emph{Proceedings of the IEEE conference on computer vision and
  pattern recognition}, pages 1646--1654, 2016.

\bibitem[Kong et~al.(2021)Kong, Zhao, Qiao, and Dong]{kong2021classsr}
Xiangtao Kong, Hengyuan Zhao, Yu~Qiao, and Chao Dong.
\newblock Classsr: A general framework to accelerate super-resolution networks
  by data characteristic.
\newblock \emph{arXiv preprint arXiv:2103.04039}, 2021.

\bibitem[Ledig et~al.(2017)Ledig, Theis, Husz{\'a}r, Caballero, Cunningham,
  Acosta, Aitken, Tejani, Totz, Wang, et~al.]{ledig2017photo}
Christian Ledig, Lucas Theis, Ferenc Husz{\'a}r, Jose Caballero, Andrew
  Cunningham, Alejandro Acosta, Andrew Aitken, Alykhan Tejani, Johannes Totz,
  Zehan Wang, et~al.
\newblock Photo-realistic single image super-resolution using a generative
  adversarial network.
\newblock In \emph{Proceedings of the IEEE conference on computer vision and
  pattern recognition}, pages 4681--4690, 2017.

\bibitem[Li et~al.(2020)Li, Zhou, Qi, Jiang, Lu, and Jia]{li2020lapar}
Wenbo Li, Kun Zhou, Lu~Qi, Nianjuan Jiang, Jiangbo Lu, and Jiaya Jia.
\newblock Lapar: Linearly-assembled pixel-adaptive regression network for
  single image super-resolution and beyond.
\newblock \emph{Advances in Neural Information Processing Systems}, 33, 2020.

\bibitem[Lim et~al.(2017)Lim, Son, Kim, Nah, and Mu~Lee]{lim2017enhanced}
Bee Lim, Sanghyun Son, Heewon Kim, Seungjun Nah, and Kyoung Mu~Lee.
\newblock Enhanced deep residual networks for single image super-resolution.
\newblock In \emph{Proceedings of the IEEE conference on computer vision and
  pattern recognition workshops}, pages 136--144, 2017.

\bibitem[Park et~al.(2020)Park, Yoo, Cho, Kim, and Kim]{park2020fast}
Seobin Park, Jinsu Yoo, Donghyeon Cho, Jiwon Kim, and Tae~Hyun Kim.
\newblock Fast adaptation to super-resolution networks via meta-learning.
\newblock \emph{arXiv preprint arXiv:2001.02905}, 5, 2020.

\bibitem[Shaham et~al.(2019)Shaham, Dekel, and Michaeli]{shaham2019singan}
Tamar~Rott Shaham, Tali Dekel, and Tomer Michaeli.
\newblock Singan: Learning a generative model from a single natural image.
\newblock In \emph{Proceedings of the IEEE/CVF International Conference on
  Computer Vision}, pages 4570--4580, 2019.

\bibitem[Shi et~al.(2016)Shi, Caballero, Husz{\'a}r, Totz, Aitken, Bishop,
  Rueckert, and Wang]{shi2016real}
Wenzhe Shi, Jose Caballero, Ferenc Husz{\'a}r, Johannes Totz, Andrew~P Aitken,
  Rob Bishop, Daniel Rueckert, and Zehan Wang.
\newblock Real-time single image and video super-resolution using an efficient
  sub-pixel convolutional neural network.
\newblock In \emph{Proceedings of the IEEE conference on computer vision and
  pattern recognition}, pages 1874--1883, 2016.

\bibitem[Shocher et~al.(2018)Shocher, Cohen, and Irani]{shocher2018zero}
Assaf Shocher, Nadav Cohen, and Michal Irani.
\newblock “zero-shot” super-resolution using deep internal learning.
\newblock In \emph{Proceedings of the IEEE Conference on Computer Vision and
  Pattern Recognition}, pages 3118--3126, 2018.

\bibitem[Shrivastava et~al.(2016)Shrivastava, Gupta, and
  Girshick]{shrivastava2016training}
Abhinav Shrivastava, Abhinav Gupta, and Ross Girshick.
\newblock Training region-based object detectors with online hard example
  mining.
\newblock In \emph{Proceedings of the IEEE conference on computer vision and
  pattern recognition}, pages 761--769, 2016.

\bibitem[Soh et~al.(2020)Soh, Cho, and Cho]{soh2020meta}
Jae~Woong Soh, Sunwoo Cho, and Nam~Ik Cho.
\newblock Meta-transfer learning for zero-shot super-resolution.
\newblock In \emph{Proceedings of the IEEE/CVF Conference on Computer Vision
  and Pattern Recognition}, pages 3516--3525, 2020.

\bibitem[Sun et~al.(2020)Sun, Chen, Slabaugh, and Torr]{sun2020learning}
Shuyang Sun, Liang Chen, Gregory Slabaugh, and Philip Torr.
\newblock Learning to sample the most useful training patches from images.
\newblock \emph{arXiv preprint arXiv:2011.12097}, 2020.

\bibitem[Xin et~al.(2020)Xin, Wang, Jiang, Li, Huang, and
  Gao]{xin2020binarized}
Jingwei Xin, Nannan Wang, Xinrui Jiang, Jie Li, Heng Huang, and Xinbo Gao.
\newblock Binarized neural network for single image super resolution.
\newblock In \emph{European Conference on Computer Vision}, pages 91--107.
  Springer, 2020.

\bibitem[Xu et~al.(2019)Xu, Ma, and Sun]{xu2019towards}
Xiangyu Xu, Yongrui Ma, and Wenxiu Sun.
\newblock Towards real scene super-resolution with raw images.
\newblock In \emph{Proceedings of the IEEE/CVF Conference on Computer Vision
  and Pattern Recognition}, pages 1723--1731, 2019.

\bibitem[Yoo et~al.(2020)Yoo, Ahn, and Sohn]{yoo2020rethinking}
Jaejun Yoo, Namhyuk Ahn, and Kyung-Ah Sohn.
\newblock Rethinking data augmentation for image super-resolution: A
  comprehensive analysis and a new strategy.
\newblock In \emph{Proceedings of the IEEE/CVF Conference on Computer Vision
  and Pattern Recognition}, pages 8375--8384, 2020.

\bibitem[Yun et~al.(2019)Yun, Han, Oh, Chun, Choe, and Yoo]{yun2019cutmix}
Sangdoo Yun, Dongyoon Han, Seong~Joon Oh, Sanghyuk Chun, Junsuk Choe, and
  Youngjoon Yoo.
\newblock Cutmix: Regularization strategy to train strong classifiers with
  localizable features.
\newblock In \emph{Proceedings of the IEEE/CVF International Conference on
  Computer Vision}, pages 6023--6032, 2019.

\bibitem[Zhang et~al.(2017)Zhang, Cisse, Dauphin, and
  Lopez-Paz]{zhang2017mixup}
Hongyi Zhang, Moustapha Cisse, Yann~N Dauphin, and David Lopez-Paz.
\newblock mixup: Beyond empirical risk minimization.
\newblock \emph{arXiv preprint arXiv:1710.09412}, 2017.

\bibitem[Zhang et~al.(2018{\natexlab{a}})Zhang, Li, Li, Wang, Zhong, and
  Fu]{zhang2018image}
Yulun Zhang, Kunpeng Li, Kai Li, Lichen Wang, Bineng Zhong, and Yun Fu.
\newblock Image super-resolution using very deep residual channel attention
  networks.
\newblock In \emph{Proceedings of the European conference on computer vision
  (ECCV)}, pages 286--301, 2018{\natexlab{a}}.

\bibitem[Zhang et~al.(2018{\natexlab{b}})Zhang, Tian, Kong, Zhong, and
  Fu]{zhang2018residual}
Yulun Zhang, Yapeng Tian, Yu~Kong, Bineng Zhong, and Yun Fu.
\newblock Residual dense network for image super-resolution.
\newblock In \emph{Proceedings of the IEEE conference on computer vision and
  pattern recognition}, pages 2472--2481, 2018{\natexlab{b}}.

\end{thebibliography}
\end{document}